\documentclass[11pt]{article}

\usepackage[final]{acl}

\usepackage{times}
\usepackage{latexsym}
\usepackage{algorithm}
\usepackage{algpseudocode}

\usepackage{bbm}
\usepackage[para]{threeparttable}
\usepackage{booktabs}
\usepackage{multirow}
\usepackage{amsmath}
\usepackage{amssymb}
\usepackage{xcolor}
\usepackage{hyperref}
\usepackage{mathtools}
\usepackage{tcolorbox}
\usepackage{listings}
\usepackage{enumitem}
\usepackage{makecell}
\definecolor{c1}{HTML}{0049C0}
\usepackage{colortbl}
\usepackage{tabularx}
\usepackage{pifont}

\usepackage[T1]{fontenc}

\usepackage[utf8]{inputenc}

\usepackage{microtype}

\usepackage{inconsolata}

\usepackage{graphicx}
\usepackage{colortbl}
\usepackage{arydshln}

\definecolor{azure(web)(azuremist)}{rgb}{0.94, 1.0, 1.0}
\definecolor{oldlace}{rgb}{0.99, 0.96, 0.9}
\definecolor{pearl}{rgb}{0.94, 0.92, 0.84}
\definecolor{seashell}{rgb}{1.0, 0.96, 0.93}
\definecolor{silver}{rgb}{0.75, 0.75, 0.75}
\definecolor{platinum}{rgb}{0.9, 0.89, 0.89}
\definecolor{almond}{rgb}{0.94, 0.87, 0.8}
\definecolor{lightskyblue}{RGB}{173, 216, 230}
\definecolor{darkgreen}{RGB}{2, 191, 35}

\newcommand{\bone}{$\text{B}_\text{1}$}
\newcommand{\btwo}{$\text{B}_\text{2}$}
\newcommand{\bthree}{$\text{B}_\text{3}$}
\newcommand{\bfour}{$\text{B}_\text{4}$}
\newcommand{\meteor}{MR}
\newcommand{\rougel}{$\text{R}_\text{L}$}

\title{Discourse Coherence and Response-Guided Context Rewriting for Multi-Party Dialogue Generation}

\author{Zhiyu Cao, Peifeng Li\thanks{ \ \ Corresponding author}, Qiaoming Zhu \\
        School of Computer Science and Technology, Soochow University, China  \\
        \texttt{zycao18@stu.suda.edu.cn}, \texttt{\{pfli, qmzhu\}@suda.edu.cn}
        }

\begin{document}
\maketitle
\begin{abstract}
Previous research on multi-party dialogue generation has predominantly leveraged structural information inherent in dialogues to directly inform the generation process. However, the prevalence of colloquial expressions and incomplete utterances in dialogues often impedes comprehension and weakens the fidelity of dialogue structure representations, which is particularly pronounced in multi-party dialogues. In this work, we propose a novel framework DRCR~(Discourse coherence and Response-guided Context Rewriting) to improve multi-party dialogue generation through dialogue context rewriting. Specifically, DRCR employs two complementary feedback signals, discourse coherence and response quality, to construct preference data for both context rewriting and response generation. Moreover, we propose a dynamic self-evolution learning method that allows the rewriter and responder to continuously enhance their capabilities through mutual interaction in an iterative training loop. Comprehensive experiments conducted on four multi-party dialogue datasets substantiate the effectiveness of DRCR.
\end{abstract}

\section{Introduction}
As a core module, dialogue generation has a significant impact on the development of dialogue systems. Despite the significant progress made in recent years with two-party dialogues (e.g., emotional support dialogues~\cite{ESConv,SUPPORTER,ES_VR} and task-oriented dialogues~\cite{InstructTODS,DivTOD,OmniDialog}), especially driven by Large Language Models (LLMs), research on Multi-party Dialogue Generation (MDG) is still scarce.

\begin{figure}[t]
\begin{center}
 \includegraphics[width=1\linewidth]{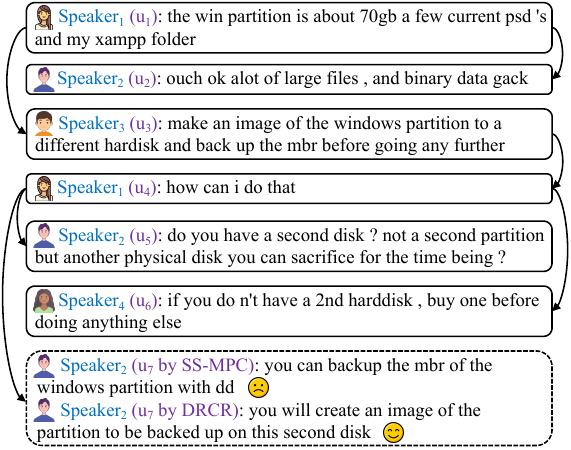}
 \caption{An example of multi-party dialogue generation. The start and end of the arrow represent the addressee and sender of an utterance, respectively.}
 \label{fig:task}
\end{center}
\vspace{-0.6cm}
\end{figure}

Unlike two-party dialogues, multi-party dialogues involve multiple roles and have a more complex structure. As shown in Figure~\ref{fig:task}, given the utterances $u_1$ to $u_6$, the goal of MDG is to generate the response $u_7$ to $u_4$. This conversation involves four speakers and has a complex discourse structure which spans multiple utterances. In contrast, the interpersonal relationship between speaker and addressee in two-party dialogue is typically manifested only across adjacent utterances, and the discourse structure is generally simpler than that of multi-party dialogue.

Previous research on MDG explicitly~\cite{MDG_data_2019_GSN,HeterMPC,SS-MPC} or implicitly~\cite{EMMDG,MADNet,RL-TRC} encodes dialogue structure information to generate responses. However, they do not take into account those informal or ambiguous expressions (such as coreference and ellipsis) in dialogues and disrupt the coherence of the dialogue discourse structure. As shown in Figure~\ref{fig:task}, the colloquial expression in $u_2$ and the reference involved with ``that'' in $u_4$ cause interference with understanding the context. This poor coherence interferes with the encoding of the dialogue structure, which in turn is detrimental to response generation. For example, the response $u_7$ generated by SS-MPC~\cite{SS-MPC} lacks continuity with the previous $u_5$, which is potentially attributed to the coreference in $u_4$ disrupting the semantics of the context. Hence, how to construct coherent context is crucial for MDG, which has been overlooked in previous research. In addition, the primary objective of context rewriting is to enhance the overall quality of dialogue generation. Therefore, the effectiveness of response generation must be explicitly integrated as a critical dimension in the rewriting process.

In this work, we propose a novel MDG method DRCR (Discourse coherence and Response-guided Context Rewriting), which is the first to incorporate context rewriting to enhance multi-party dialogue generation. DRCR introduces a dual-feedback framework that leverages discourse coherence quality and response quality as optimization signals to guide the rewriting process. To enhance the understanding of dialogue context, DRCR achieves two complementary objectives: (1) clarifying the underlying dialogue structure, and (2) facilitating the generation of contextually aligned and semantically appropriate responses.

Specifically, we employ LLMs to play the roles of rewriter and responder for dialogue context rewriting and response generation, respectively. To initialize the behavior of the rewriter and the responder, DRCR first performs warm-up using preference data generated by an external LLM, and then updates the preference data through mutual self-evolution to adapt to the model's alignment state. Experimental results on four datasets demonstrate the effectiveness of DRCR. In summary, our contributions are as follows.
\begin{itemize}
\vspace{-5pt}
\item We propose a novel method DRCR which incorporates context rewriting to enhance multi-party dialogue generation.
\vspace{-8pt}
 \item DRCR introduces discourse coherence and response quality as dual feedback signals to guide context rewriting.
\vspace{-8pt}
\item To reduce dependence on external preference data and mitigate biases arising from static alignment, we employ mutual self-evolution between the rewriter and the responder.
\end{itemize}

\section{Related Work}
Previous work on MDG falls into two paradigms: explicit~\cite{MDG_data_2019_GSN,HeterMPC,SS-MPC} and implicit~\cite{EMMDG,MADNet,RL-TRC} incorporation of dialogue structure information. The former used graph networks for dialogue modeling, while the latter integrated dialogue structure information through latent variable prediction or reinforcement learning.

GSN~\cite{MDG_data_2019_GSN} used an utterance-level graph-structured encoder for dialogue modeling. Since the responses generated in MDG depend on the interlocutor and the historical utterances, HeterMPC~\cite{HeterMPC} proposed a heterogeneous graph-based neural network that used two node types to model utterances and interlocutors. Considering that using graph-based methods may cause information loss in the process of mapping utterances to structural embeddings, SS-MPC~\cite{SS-MPC} introduced sequential input to convert dialogue structures.

Unlike the methods that explicitly incorporate dialogue structure information, some studies regarded the perception of dialogue structure as an optimization objective. EMMDG~\cite{EMMDG} used the expectation-maximization-based method to iteratively generate addressee labels through the expectation steps and optimized the response generation model through the maximization steps. To ensure message passing between dialogue fragments, MADNet~\cite{MADNet} designed four latent edges and proposed an EM-based method to optimize the edge-type-dependent message passing. RL-TRC~\cite{RL-TRC} introduced reinforcement learning to guide the generated responses to remain consistent with the target utterances in terms of topic and logic.

In contrast to the above methods, our DRCR enhances the coherence of dialogue structure through context rewriting, thereby enabling the model to achieve enhanced perception of dialogue structure.

\begin{figure*}[t]
\begin{center}
 \includegraphics[width=1\linewidth]{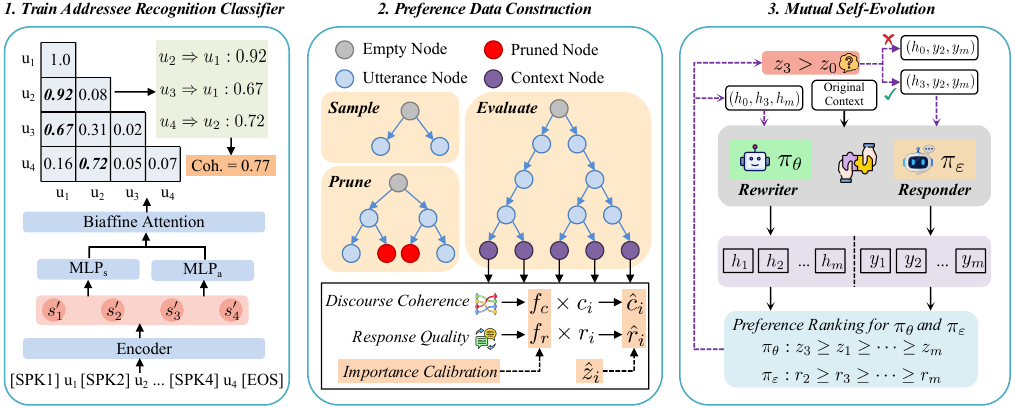}
 \vspace{-14pt}
 \caption{An overview of DRCR, which includes three stages: 1) Train Addressee Recognition Classifier: Identifying the addressee corresponding to each utterance in preparation for scoring the discourse coherence of the dialogue context, 2) Preference Data Construction: Constructing preference data based on the discourse coherence of the dialogue and the quality of responses, and 3) Mutual Self-Evolution: Iteratively optimizing both rewriter and responder without relying on external data.}
 \label{fig:method}
\end{center}
\vspace{-12pt}
\end{figure*}

\section{Methodology}

Given a multi-party dialogue context $H_t=\{(s_1,a_1,u_1),(s_2,a_2,u_2),\dots,(s_{t},a_{t},u_{t})\}$ comprising $t$ turns, where $s_i$, $a_i$, and $u_i$ denote the speaker identifier, the addressee identifier (i.e., who this utterance is addressed to), and the utterance of the $i^{th}$ turn, respectively, the objective of MDG is to generate the response $u_{t+1}$ for the $(t+1)^{th}$ turn conditioned on $H_t$ and the current speaker and addressee $(s_{t+1},a_{t+1})$ as follows, 
\begin{equation}
\small
    u_{t+1}=\mathop{argmax} \limits_{y} \sum_{k=1}^{|y|} logP(y_k|H_t,s_{t+1},a_{t+1},y_{<k}), 
\end{equation}
where $y_k$ and $y_{<k}$ are the $k^{th}$ token and the preceding $(k-1)$ tokens of the response $y$, respectively.

\subsection{Overview}
As shown in Figure~\ref{fig:method}, the DRCR framework operates in three stages. In the first stage, an utterance addressee classifier is trained to identify the addressee of each utterance (\autoref{DDCR}). Based on the predicted probabilities of this classifier, in the second stage, we use discourse coherence quality and response generation quality to evaluate the quality of sampled candidate rewritten contexts. The scores for these two dimensions are calibrated using the Coefficient of Variation, and the highest- and lowest-scoring rewritten contexts and responses are selected as preference data to initialize the rewriter and responder (\autoref{PDC}). Finally, the third stage establishes an iterative self-optimization loop where the rewriter and responder collaborate to build new preference data for ongoing optimization (\autoref{MSE}).

\subsection{Addressee Recognition}
\label{DDCR}
Inspired by previous research~\cite{Coherence_Modeling} showing that discourse relations and coherence are closely related, we adopt Addressee Recognition (AR) on dialogue context as a proxy for discourse coherence prediction. The underlying rationale is that the clarity and logical flow of a conversation are reflected in the ease with which an addressee can be identified from the surrounding discourse. If an AR model assigns high confidence to the correct addressee, it suggests that the dialogue exhibits strong coherence. Therefore, the predicted probability assigned by the classifier to the true addressee serves as a quantifiable indicator of discourse coherence.

To assess the coherence of dialogue discourse and thereby evaluate the quality of various rewritten dialogue contexts, we first trained an AR classifier. Specifically, we sequentially concatenate the speaker's corresponding tags and utterances to obtain the dialogue context $x=s_1 \oplus u_1 \oplus s_2 \oplus u_2 \dots \oplus s_n \oplus u_n$, and then use \texttt{roberta-large} as an encoder $Enc(\cdot)$ to encode the dialogue context:
\begin{equation}
    Enc(x) = s_1^{\prime} \oplus u_1^{\prime} \oplus s_2^{\prime} \oplus u_2^{\prime} \dots \oplus s_n^{\prime} \oplus u_n^{\prime},
\end{equation}
where $s_i^{\prime}$ and $u_i^{\prime}$ represent the hidden representations corresponding to $s_i$ and $u_i$, respectively. We take $\mathcal{S} = \{s_1^{\prime}, \dots, s_n^{\prime}\}$ as the representation of the context. Next, we introduce two multi-layer perceptrons $\text{MLP}_s$ and $\text{MLP}_a$ to map $\mathcal{S}$ into the representations of senders and addressees, and then use biaffine attention~\cite{Biaffine_Attention} to model pairwise associations between utterances:
\begin{equation}
\begin{split}
    &v_{i}^s = \text{MLP}_s(s_i^{\prime}),v_{i}^a = \text{MLP}_a(s_i^{\prime}),  \\
    &\mathrm{score}(i \leftarrow j)=\left[\begin{array}{c}
v_{i}^s \\
1
\end{array}\right]^{\mathrm{T}} W^b v_{j}^a, 
\end{split}
\end{equation}
where $W^b$ is the biaffine parameter, $\mathrm{score}(i \leftarrow j)$ represents the score of $u_j$ as the addressee of $u_i$. It is worth noting that all scores can be efficiently calculated simultaneously in matrix form.

The optimization objective of the AR classifier is to minimize the cross-entropy loss function: 
\begin{equation}
\small
L_{c}=- \frac{1}{n} \sum_{1 \leq i \leq n} \log \frac{e^{\mathrm{score}(i \leftarrow a_i)}}{\sum_{1 \leq k < i, k \neq a_i} e^{\mathrm{score}(i \leftarrow k)}}.
\end{equation}
To explore the impact of addressee recognition, we discussed it in Section~\ref{sec:alternative_coh} and Appendix~\ref{sec:Addressee_Recognition_Module}.

\subsection{Preference Data Construction}
\label{PDC}
Since high-quality dialogue context is crucial for response generation, we design a rewriter and a responder specifically for context rewriting and response generation. To train them, we construct preference data for context rewriting and response generation through an external LLM $\pi_{t}$ as a teacher rewriter to perform preference alignment. We propose a tree-based top-down sampling method to construct diverse candidate rewritten contexts. This approach promotes diversity by progressively exploring multiple rewriting paths at each dialogue turn, preserves semantic fidelity by pruning utterances that deviate from the original meaning using Natural Language Inference (NLI)-based filtering, and facilitates exploration of potential contextual improvements across the entire conversation.

Starting from the root node of the first layer, which is an empty node representing an empty context, we use $\pi_{t}$ as the rewriting model to sample multiple candidate rewritten utterances for each utterance one by one (Appendix~\ref{sec:utterance_level_rewriting} for the prompt). The nodes in the $(i+1)$-th layer represent the candidate rewritten utterances corresponding to the $i$-th utterance in the dialogue context, and a path from the root node to a leaf node constitutes a candidate rewritten context. Assuming the currently sampled partial context is $\hat{h}_{l-1}=\{\hat{u}_1,\hat{u}_2,\dots,\hat{u}_{l-1}\}$, which is a path from 2-nd layer to $(l+1)$-th layer, then we use $\pi_{t}$ to sample $n$ candidate rewritten utterances for the $l$-th utterance:
\begin{equation}
    \{\hat{u}_l^i\}_{i=1}^n \sim P_{LLM}(\{\hat{u}_1,\hat{u}_2,\dots,\hat{u}_{l-1}\}).
\end{equation}

These sampled rewritten utterances are the child nodes of $\hat{u}_{l-1}$. To prevent the generated candidates from deviating from the original utterance, we measure the NLI score between the original utterance and the rewritten utterance and prune candidates whose NLI score is below $\lambda$ (i.e., red nodes in Figure~\ref{fig:method}). After filtering, each candidate $\hat{u}_l^i$ will be appended to $\hat{h}_{l-1}$ to form $\hat{h}_l^i={\hat{h}_{l-1} \cup \hat{u}_l^i}$, the rewriting of the next utterance continues until the entire dialogue context is rewritten.

Through the above sampling, we can obtain $m$ candidate rewritten contexts $\mathcal{H}=\{h_{i}\}_m$ (i.e., context nodes in Figure~\ref{fig:method}). To evaluate the quality of the candidates, we score them along two dimensions: discourse coherence of context and quality of responses generated based on dialogue context.

\noindent \textbf{Discourse Coherence} Dialogue discourse coherence is the foundation for understanding dialogue context. More coherent dialogue is more conducive to generating reasonable responses. To measure discourse coherence, we first employ the classifier trained in Section~\ref{DDCR} to identify the addressee corresponding to each utterance. Our insight is that for each dialogue discourse, if the classifier can accurately identify the addressee, it indicates that the dialogue discourse is more coherent. Therefore, we take the average of the predicted probabilities of the true addressee corresponding to each utterance as the coherence score $\mathcal{C}=\{c_i\}_m$: 
\begin{equation}
\vspace{-0.1cm}
\label{coherence}
    c_i=\frac{1}{t-1}\sum_{i=2}^t \mathrm{score}(i \leftarrow a_i).
    \vspace{-0.1cm}
\end{equation}
\textbf{Response Quality} Response quality serves as more direct and intuitive feedback on context rewriting. Higher-quality dialogue contexts facilitate the generation of correspondingly higher-quality responses. We use a responder  model $\pi_\varepsilon$ to generate the response $y_i$ for each candidate $h_{i}$ and compare $\mathcal{Y}=\{y_{i}\}_m$ with the ground truth as a proxy $\mathcal{R}=\{r_i\}_m$ for evaluating the quality of $h_{i}$\footnote{We also tried adding $\text{BLEU}_\text{4}$ and METEOR for evaluation, but did not observe any significant performance improvement.}: 
\begin{equation}
\vspace{-0.1cm}
\label{response}
    r_i=\frac{1}{2}(\text{BLEU}_1(y_{i},y_{i}^{gt})+\text{ROUGE}_l(y_{i},y_{i}^{gt})), 
    \vspace{-0.1cm}
\end{equation}
where $y_{i}^{gt}$ represents the ground truth response corresponding to $h_{i}$. $\text{BLEU}_1$ and $\text{ROUGE}_l$ measure the similarity between $y_{i}$ and $y_{i}^{gt}$ from the perspectives of precision and recall, respectively.

\noindent \textbf{Coefficient of Variation Guided Importance Calibration} By integrating discourse coherence $c_i$ with response quality $r_i$, both the coherence of discourse structure and the benefits for response generation can be considered simultaneously. However, the issue with this approach is that different samples require focusing on different aspects. For example, when the discourse coherence $c_i$  is already strong enough, greater emphasis should be placed on response quality $r_i$, and vice versa.

To address the above issues, we introduce the Coefficient of Variation to calculate the weights of $c_i$ and $r_i$. Given discourse coherence $\{c_i\}_m$, we calculate its corresponding standard deviation $\sigma_c$ and mean $\mu_c$, and the same applies to response quality $\{r_i\}_m$ (i.e., $\sigma_r$ and $\mu_r$). Then the corresponding weight coefficients of $c_i$ and $r_i$ are $f_c$ and $f_r$: 
\begin{equation}
\vspace{-0.2cm}
    f_c=\frac{\sigma_c}{\mu_c},f_r=\frac{\sigma_r}{\mu_r}.   
\end{equation}

It is worth noting that the values of $f_c$ and $f_r$ are independent of magnitude, so even a very small magnitude can be influential if it is variable. Then we use the softmax function for normalization: 
\begin{equation}
    \alpha_c=\frac{e^{f_c/\tau}}{e^{f_c/\tau}+e^{f_r/\tau}},\alpha_r=\frac{e^{f_r/\tau}}{e^{f_c/\tau}+e^{f_r/\tau}}, 
\end{equation}
where temperature coefficient $\tau$ controls the smoothness of weights to avoid abnormal weights. According to the calculated weights, we combine $c_i$ and $r_i$ to derive the rewriting score $\mathcal{Z}=\{z_i\}_m$:
\begin{equation}
\label{ovs}
    z_i=\alpha_c \cdot c_i+\alpha_r \cdot r_i.
\end{equation}
\noindent \textbf{Warm-up Preference Alignment} For response generation, we only need to consider feedback on the response quality, while for context rewriting, we need to further include additional feedback on discourse coherence. Therefore, the candidate rewritten contexts are ranked according to the rewriting score $z_i$ and response scores $r_i$, respectively. The samples with the highest and lowest scores are designated as the chosen and rejected samples, respectively. This process yields the preference datasets for context rewriting ($D_h$) and response generation ($D_r$): 
\begin{equation*}
\label{PD}
\small
\begin{split}
    &D_h \sim \{(H_t,h_w,h_l)|w=\text{argmax}(\mathcal{Z}),l=\text{argmin}(\mathcal{Z})\}, \\
    &D_r \sim \{(H_t,y_w,y_l)|w=\text{argmax}(\mathcal{R}),l=\text{argmin}(\mathcal{R})\}, 
\end{split}
\end{equation*}
where $H_t$ is the original dialogue context, $h_w$ and $h_l$ are the preferred and dispreferred rewritten contexts, respectively. After preparing the preference data $D_h$ and $D_r$, we employ DPO~\cite{DPO} to train the rewriter $\pi_\theta$ and the responder $\pi_\varepsilon$, respectively. The optimization objectives for rewriter and responder are $\mathcal{L}_\theta$ and $\mathcal{L}_\varepsilon$, respectively: 
\begin{equation}
\small
\begin{aligned}
\mathcal{R}_\theta &(H,h)=\beta \, \text{log}\frac{\pi_{\theta}(h|H)}{\pi_\theta^{\text{ref}}(h|H)}, \mathcal{R}_\varepsilon (H,y)=\beta \, \text{log}\frac{\pi_{\varepsilon}(y|H)}{\pi_\varepsilon^{\text{ref}}(y|H)},\\
\mathcal{L}_\theta&=-\mathbb{E}_{(H,h_w,h_l)\sim \mathcal{D}_{h}}\text{log}\, \sigma \left (\mathcal{R}_\theta(H,h_w)-\mathcal{R}_\theta(H,h_l)\right ), \\
\mathcal{L}_\varepsilon&=-\mathbb{E}_{(H,y_w,y_l)\sim \mathcal{D}_{r}}\text{log}\, \sigma \left (\mathcal{R}_\varepsilon(H,y_w)-\mathcal{R}_\varepsilon(H,y_l)\right ), \\
\end{aligned}
\end{equation}
where $\sigma$ represents the sigmoid function, $\pi_\theta^{\text{ref}}$ and $\pi_\varepsilon^{\text{ref}}$ are the reference policies for $\pi_\theta$ and $\pi_\varepsilon$ before performing DPO.

\begin{algorithm}[t]
\small
    \caption{Mutual Self-Evolution}
    \label{alg:mse}
    \vspace{-1mm}
\begin{algorithmic}[1]
    \State \textbf{Input: } Rewriter $\pi_\theta$,  Responder $\pi_\varepsilon$, Iterations $T$, Rewriting threshold $\phi$, MDG dataset $\mathcal{D}$
    \For{$k=1$ {\bfseries to} $T$}
    \State For each $(H_t,s_{t+1},a_{t+1}) \in \mathcal{D}$, sample $m$ candidate rewritten contexts $\{h_i^k\}_m \sim \pi_\theta(H_t)$
    \State Generate response $\{y_i^k\}_m \sim \pi_\varepsilon(\{h_i^k\}_m)$
    \State Calculate $\mathcal{C}_k=\{c_i^k\}_m$ and $\mathcal{R}_k=\{r_i^k\}_m$ according to Eq. (\ref{coherence}) and Eq. (\ref{response}).
    \State Calculate $\mathcal{Z}_k=\{z_i^k\}_m$ according to Eq. (\ref{ovs}).
    \State Calculate the rewriting score $z_0$ for $H_t$.
    \If {$\operatorname{max}(\{z_i^k\}_m)>z_0$}
        \State $H_t \gets \operatorname{argmax}_{h_i} (\{z_i^k\}_m)$
    \EndIf
    \State Collect preference datasets $D_h^{k}$ and $D_r^{k}$ based on $\mathcal{Z}_k$ and $\mathcal{R}_k$.
    \State Update $\pi_\theta$ using $\mathcal{L}_\theta$ on $D_h^{k}$.
    \State Update $\pi_\varepsilon$ using $\mathcal{L}_\varepsilon$ on $D_r^{k}$.
    \EndFor
    \State \Return $\pi_\theta,\pi_\varepsilon$ 
\end{algorithmic}
\end{algorithm}

\subsection{Mutual Self-Evolution}
\label{MSE}
Although the preference data we constructed can endow the model with the ability to rewrite and respond, this static training mode heavily relies on externally constructed preference data. Moreover, the preferences of the rewriter and responder may evolve during training, yet static preference data cannot accommodate such dynamic shifts in the models' alignment. Hence, we propose iteratively optimizing the rewriter and responder through mutual self-evolution, enabling them to reduce dependence on external data and update preference data based on their alignment status.

The specific mutual self-evolution process refers to Algorithm~\ref{alg:mse}. In the $k$-th round of mutual self-evolution iteration, we use the rewriter $\pi_\theta$ from the previous iteration to sample $m$ rewritten candidates $\{h_i^k\}_m$ for the dialogue context $H_t$, and use the responder $\pi_\varepsilon$ to generate a response for each candidate to obtain $\{y_i^k\}_m$. Specifically, the rewriter and responder of the first iteration are models initialized with preference data from Section~\ref{PDC}. Then we score each candidate based on discourse coherence and response quality, obtaining the rewriting score $\mathcal{Z}_k=\{z_i^k\}_m$ and response score $\mathcal{R}_k=\{r_i^k\}_m$.

To alleviate the noise interference caused by informal and incomplete dialogue context in response generation, low-quality dialogue context will be replaced with rewritten context (Lines 8-10 of Algorithm~\ref{alg:mse}). In the iterative process of mutual self-evolution, we analyze the rewriting score $z_0$ of the original context and the most preferred candidate context $h_j=\text{argmax}_{h_i} (\{z_i^k\}_m)$. If $z_j^k>z_0$, then we replace the original context with $h_j$.

Finally, DRCR selects the highest and lowest scoring samples as the chosen and rejected samples according to Eq. (\ref{ovs}) based on the rewriting score $\mathcal{Z}_k$ and the response score $\mathcal{R}_k$, resulting in the context rewriting preference optimization data $D_h^{k}$ and the response preference optimization data $D_r^{k}$, which are used to train the rewriter and responder, respectively. After each iteration, the procedure is repeated to iteratively refine both models.

\begin{table*}[t]
    \centering
    \footnotesize
    \resizebox{\textwidth}{!}
    {\begin{tabular}{lcccccccccccc}
    \toprule
        \multirow{2}{*}{\bf Method} & \multicolumn{6}{c}{\bf Ubuntu IRC-16} & \multicolumn{6}{c}{\bf Ubuntu IRC-19}  \\
        \cmidrule(lr){2-7}\cmidrule(lr){8-13}
         & \bone & \btwo & \bthree  & \bfour & \meteor & \rougel & \bone & \btwo & \bthree  & \bfour & \meteor & \rougel \\
        \hline
        \rowcolor{pearl}
        \multicolumn{13}{c}{\textbf{\textsl{SLM-based MDG}}} \\
        \hline
        {\bf GSN}~\cite{MDG_data_2019_GSN} &   6.32 & 2.28 & 1.10 & 0.61 & 3.27 & 7.39  &   10.23 & 3.57 & 1.70 & 0.97 & 4.10 & 9.91 \\
        {\bf HeterMPC}~\cite{HeterMPC} &  11.40 & 4.29 & 2.43 & 1.74 & 4.57 & 10.44  &  12.26 & 4.80 & 2.42 & 1.49 & 4.94 & 11.20 \\
        {\bf EMMDG}~\cite{EMMDG} &  11.67 & 4.73 &  2.64 &  1.81 &  5.12 &  10.43  &  12.31 &  5.39 &  3.34 &  2.45 &  5.52 &  11.71 \\
        {\bf MADNet}~\cite{MADNet} &  11.82 & 4.58 & 2.65 & 1.91 & 4.90 & 10.74  &  12.73 & 5.12 & 2.64 & 1.63 & 5.31 & 11.74 \\
        {\bf RL-TRC}~\cite{RL-TRC} &  12.52 & 5.41 & 3.34 &  2.45 & 5.45 & 11.31  &  13.66 & 6.58 & 4.10 & 2.93 & 6.20 & 12.72 \\
        {\bf SS-MPC}~\cite{SS-MPC} &  13.40 &  5.87 &  3.60 &  2.65 &  6.97 & 11.14  &  15.60 &  6.62 &  3.67 &  2.44 &  7.75 &  12.44\\
        \hline
        \rowcolor{pearl}
        \multicolumn{13}{c}{\textbf{\textsl{LLM-based MDG}}} \\
        {\bf Llama3.2-3B+SFT} & 12.02 & 4.75 & 2.78 & 2.03 & 5.01 & 11.25 & 13.23 & 5.87 & 3.37 & 2.20 & 6.02 & 11.94 \\
        {\bf Llama3.2-3B+DRCR (Ours)} & 14.51 & 6.23 & 4.12 & 2.97 & 7.64 & 12.27 & 16.91 & 7.24 & 4.95 & 3.71 & 7.92 & 13.37\\
        \hdashline
        {\bf Qwen3-4B+SFT} & 12.34 & 5.07 & 2.94 & 2.19 & 5.18 & 11.22 &  13.48 & 6.05 & 3.54 & 2.39 & 6.14 & 12.20 \\
        {\bf Qwen3-4B+DRCR (Ours)} &  14.93 & 6.51 & 4.45 & 3.18 & 7.89 & 12.41 &  16.95 & 7.51 & 5.32 & 3.94 & 8.03 & 13.55 \\
        \hdashline
        {\bf Qwen3-8B+SFT} & 13.07 & 5.38 & 3.24 & 2.41 & 5.39 & 11.53 & 14.22 & 6.53 & 3.83 & 2.52 & 6.37 & 12.92 \\
        {\bf Qwen3-8B+DRCR (Ours)} & \textbf{16.04} & \textbf{7.19} & \textbf{5.27} & \textbf{4.03} & \textbf{8.16} & \textbf{13.32} &  \textbf{17.81} & \textbf{8.02} & \textbf{5.79} & \textbf{4.26} & \textbf{8.49} & \textbf{14.82} \\
    \bottomrule
    \end{tabular}}
    \caption{Experimental results on IRC-16 and IRC-19 (Section~\ref{sec:eval_HLA-Chat++_Friends} for the results on HLA-Chat++ and Friends).}
\label{table:main_exp}
\end{table*}
During the inference phase, if the discourse coherence score of a sample falls below the threshold $\phi$, it will be rewritten before generating a response. Otherwise, a response is generated directly.

\section{Experiments}
\subsection{Experimental Settings}
\noindent \textbf{Datasets} Following previous research~\cite{RL-TRC}, we evaluated our proposed DRCR on two MDG datasets, namely Ubuntu IRC-16~\cite{MDG_data_2016} and IRC-19~\cite{MDG_data_2019_GSN}. To verify the generalizability of DRCR, we also conducted evaluations on the other two MDG datasets, HLA-Chat++ and Friends in Section~\ref{sec:eval_HLA-Chat++_Friends}.

\noindent \textbf{Evaluation Metrics} Following previous work \cite{RL-TRC}, we used $\text{BLEU}_\text{1}$-$\text{BLEU}_\text{4}$  (\bone-\bfour), METEOR (\meteor), and $\text{ROUGE}_\text{L}$ (\rougel) as evaluation metrics. In addition to these automated metrics, we also conduct human evaluations from four dimensions: Coherence between the generated response and the context, as well as the Fluency, Informativeness, and Helpfulness of the generated response.

\noindent \textbf{Baselines} To verify the effectiveness of DRCR, we compared it with advanced baselines, including GSN~\cite{MDG_data_2019_GSN}, HeterMPS~\cite{HeterMPC}, EMMDG~\cite{EMMDG}, MADNet~\cite{MADNet}, RL-TRC~\cite{RL-TRC}, and SS-MPC~\cite{SS-MPC}. Considering the strong instruction-following capability of LLMs, we also compared DRCR with \texttt{Llama-3.2-3B-Instruct}, \texttt{Qwen3-4B}, and \texttt{Qwen3-8B} that are fine-tuned on the original dataset.

\noindent \textbf{Implementation Details} We evaluated DRCR using \texttt{Llama-3.2-3B-Instruct}, \texttt{Qwen3-4B}, and \texttt{Qwen3-8B} as backbone models (for both Rewriter $\pi_{\theta}$ and Responder $\pi_{\varepsilon}$) to verify its generalization, respectively. We use \texttt{gpt-3.5-turbo-0125} as the external model $\pi_t$ for constructing preference data in the warm-up phase. For the training of the addressee classifier, the number of epochs is set to 10, and the batch size is 128. In candidate rewritten utterances sampling, we use \texttt{all-mpnet-base-v2} to measure the NLI similarity between the original utterance and the rewritten utterance. The number of candidate utterance samples was set to $n = 2$ to reduce computational overhead. Further implementation details are provided in Appendix~\ref{sec:implementation}.

\subsection{Main Results}
We conducted validation on different model families (\texttt{Llama} vs. \texttt{Qwen}) and model scales (\texttt{Qwen3-4B} vs. \texttt{Qwen3-8B}) by comparing the performance of different models undergoing standard supervised fine-tuning on the original dataset (e.g., ``Qwen3-8B+SFT'') and equipped with DRCR (e.g., ``Qwen3-8B+DRCR''). As shown in Table~\ref{table:main_exp}, our DRCR significantly improves the performance on all metrics. We noticed that the three evaluated backbones were inferior to the previous SOTA models when only supervised fine-tuning was performed.  After using DRCR for preference alignment, the performance of the three backbones greatly surpasses them. This indicates that DRCR effectively improves the model's understanding of context by guiding context rewriting from the perspectives of discourse coherence and response quality, leading to more reasonable responses.

The effectiveness across different model families and scales also indicates that DRCR can be generalized to different backbones. This is attributed to the fact that DRCR is model-agnostic and can be adapted to different LLMs.

\begin{table}[t]
    \small
    \centering
    \resizebox{\linewidth}{!}{
    \begin{tabular}{lcccccc}
        \toprule
        {\textbf{Variant}} &  \bone & \btwo & \bthree & \bfour & \meteor & \rougel \\        
        \midrule
        \cellcolor[gray]{0.9}\textit{DRCR} & \cellcolor[gray]{0.9}17.81 & \cellcolor[gray]{0.9}8.02 & \cellcolor[gray]{0.9}5.79  & \cellcolor[gray]{0.9}4.26 & \cellcolor[gray]{0.9}8.49 & \cellcolor[gray]{0.9}14.82\\
        
        \quad w/o. DC & 17.09 & 7.22 & 5.13 & 3.54 & 7.75 & 13.94 \\
        \quad w/o. RQ & 16.98 & 7.16 & 5.21 & 3.57 & 7.64 & 14.03 \\
        \quad w/o. CoV & 17.18 & 7.36 & 5.27 & 3.62 & 7.81 & 14.11 \\
        \quad w/o. Warm-up & 15.17 & 6.72 & 4.26 & 3.03 & 6.84 & 13.32 \\
        \quad w/o. MSE & 15.87 & 7.24 & 4.62 & 3.37 & 7.12 & 13.77 \\
        
        \midrule
        DRCR$_\texttt{GPT-4.1}$  & 17.97 & 8.16 & 5.87 & 4.37 & 8.62 & 14.93  \\
        DRCR$_\texttt{Qwen3-8B}$  & 17.02 & 7.12 & 5.14 & 3.47 & 7.52 & 14.03 \\
        
        \bottomrule
    \end{tabular}
    }
    \caption{
    Ablation experiments (\texttt{Qwen3-8B} as backbone) and different LLMs as teacher rewriter $\pi_t$ on  IRC-19.
    }
    \label{table:ablation}
    \vspace{-11mm}
\end{table}

\subsection{Ablation Study}

\noindent \textbf{Preference Data Construction} In the preference alignment of the rewriter, we designed scoring for the two dimensions of discourse coherence (DC) and response quality (RQ). As shown in Table~\ref{table:ablation}, we removed these two items separately (``w/o. DC'' and ``w/o. RQ''). The results demonstrate that eliminating any one dimension results in performance degradation, indicating that feedback from both discourse coherence and response quality is essential for effective context rewriting. Considering the different importance of the two dimensions, we use the Coefficient of Variation (CoV) to calibrate their weights. When the Coefficient of Variation was removed (``w/o. CoV''), $\text{B}_\text{1}$ and $\text{R}_\text{L}$ decreased by 0.63 and 0.71, respectively. This indicates that the Coefficient of Variation plays a crucial role in calibrating the importance of the two dimensions. We also evaluated accuracy as the proxy in Appendix~\ref{sec:Coherence_Assessment}. The results indicate that accuracy is less effective than probabilistic measures for this purpose.

\noindent \textbf{Preference Alignment of Rewriter and Responder} The training of the rewriter and responder in DRCR is divided into two steps: warm-up preference alignment (Warm-up) and mutual self-evolution (MSE). Ablation studies confirm the necessity of both modules, as the removal of either results in performance degradation, underscoring their individual efficacy in enhancing the model's capabilities. Additionally, we also observed that removing the warm-up preference alignment resulted in a greater performance decline. This is potentially because the warm-up phase initializes the behavior of the rewriter and responder. Without warm-up preference alignment, the model's mutual self-evolution capability is weaker, and it damages the quality of preference data during the iterations.

\noindent \textbf{Model Size} In the construction of initial preference data, we used \texttt{gpt-3.5-turbo-0125} to sample candidate rewritten contexts. We analyzed the effects of sampling using different models (e.g., \texttt{GPT-4.1} and \texttt{Qwen3-8B}). As shown in Table~\ref{table:ablation}, it can be observed that preference data constructed by stronger models can more effectively improve model performance. Appendix~\ref{sec:teacher_models_data_con} provides analysis for data construction using more other LLMs.

\begin{table}[t]
      \centering
      \resizebox{\linewidth}{!}{
      \begin{tabular}{cccc}
      \toprule
        Datasets               &   Origin   &   w/ Warm-up   &  DRCR    \\
      \midrule
        Ubuntu IRC-16          &  75.32  &   81.59  &  84.16  \\
        Ubuntu IRC-19         &  84.16  &    87.30  &   91.37 \\
      \bottomrule
      \end{tabular}
      }
      \caption{Using \texttt{Qwen3-8B} as the backbone to analyze the impact of rewriting on the discourse coherence.}
      \vspace{-5mm}
      \label{tab_rewrite_analysis}
\end{table}

\subsection{Analysis}
\noindent \textbf{Impact of Rewriting on Discourse Coherence} The insight of DRCR lies in enhancing the discourse coherence through context rewriting. To verify whether rewriting can enhance discourse coherence, we used the rewriter $\pi_\theta$ to conduct validation on Ubuntu IRC-16 and IRC-19. We compared the changes in the accuracy of addressee recognition within the dialogue context before and after rewriting. As shown in Table~\ref{tab_rewrite_analysis}, we analyzed the discourse coherence of using the original context (``Origin''), as well as the discourse coherence of the rewritten context after using models that were only warmed up with preference data from Section~\ref{PDC} (``w/ Warmup'') and the final model (``DRCR''). It can be observed that after warm-up and mutual self-evolution, the coherence of the rewritten context is better than that of the original context. This indicates that rewriting can effectively enhance discourse coherence, and both stages improve the performance of the rewriter. We provide further analysis on the necessity of context rewriting in Appendix~\ref{context_rewriting_necessity}.

\begin{table}[t]
      \centering
      \resizebox{\linewidth}{!}{
      \begin{tabular}{lcc}
      \toprule
        Method               &   \bf Ubuntu IRC-16 &  \bf Ubuntu IRC-19   \\
      \midrule
        MADNet         &   72.32   &   81.79   \\
        RL-TRC         &    76.49  &   85.72   \\
        SS-MPC         &   76.93   &  86.51    \\
        Qwen3-8B+SFT         &   74.07   &   84.26  \\
        Qwen3-8B+DRCR         &   \textbf{78.96}   &  \textbf{89.34}   \\
        \midrule
        Ground Truth           &   85.67   &     96.42     \\
      \bottomrule
      \end{tabular}
      }
      \caption{Accuracy of identifying the addressee based on the generated response.}
      \vspace{-4mm}
      \label{tab_coher_eval}
\end{table}

\begin{table}[t]
    \centering
    \resizebox{\linewidth}{!}{
    \begin{tabular}{lcccccc}
        \toprule
        \multicolumn{1}{c}{\textbf{Method}} &  \bone & \btwo & \bthree & \bfour & \meteor & \rougel  \\        
        \midrule
        DRCR-LLM	& 17.75	& 8.06	& 5.82	& 4.21	& 8.44	& 14.57  \\
        DRCR-AR	& 17.81	& 8.02	& 5.79	& 4.26	& 8.49	& 14.82  \\
        \bottomrule
    \end{tabular}
    }
    \caption{
    Comparison of using LLM for coherence scoring and addressee recognition classifier for coherence measurement on Ubuntu IRC-19.
    }
    \label{table:alternative_coh}
   \vspace{-5mm}
\end{table}

\noindent \textbf{Coherence between Generated Response and Target Utterance} To further analyze whether the generated responses can also maintain coherence with the context, we use the classifier trained in Section~\ref{DDCR} to determine the coherence between the response and the target utterance, and measure it using accuracy. As shown in Table~\ref{tab_coher_eval}, the responses generated by DRCR have better coherence with the dialogue context, achieving accuracies of 78.96 and 89.34 on Ubuntu IRC-16 and IRC-19, respectively. This is attributed to the fact that DRCR can effectively enhance contextual understanding.

Additionally, we analyzed the impact of scaling the number of iterations and the number of candidates in the mutual self-evolution process on model performance, respectively in Appendices~\ref{sec:scale_iterations} and~\ref{sec:scale_candidates}, and analyzed the rewriting threshold $\phi$ in Appendix~\ref{sec:rewriting_threshold}. In Appendix~\ref{sec:time_complexity_resource_usage}, we analyze the time complexity and resource usage at each stage.

\subsection{Alternative to Discourse Coherence Measurement}
\label{sec:alternative_coh}
In this paper we introduce addressee recognition as a quantitative indicator of discourse coherence. Another candidate approach is to give a dialogue history and let LLM score its coherence. However, this approach both increases computational resource consumption and is subject to the bias of LLM preferences.

To verify the effectiveness of this method, we use LLM for coherence scoring to construct preference data. Specifically, we first introduce GPT-4 to evaluate the coherence of each dialogue history and provide a score between 1 and 5. Next we normalize these scores by batch to between 0 and 1. The results are shown in Table~\ref{table:alternative_coh} using Qwen3-8B as the backbone for analysis on Ubuntu IRC-19. ``DRCR-LLM'' indicates the performance is evaluated using LLM for coherence scoring, while ``DRCR-AR'' represents the coherence is measured by the probability of the addressee classifier.

It can be observed that constructing preference data through LLM scoring performs almost on par with using an addressee classifier for evaluation, and is even worse in terms of ROUGE. However, the computational overhead of evaluation through the classifier is much lower than scoring with LLM.

\subsection{Evaluation on HLA-Chat++ and Friends}
\label{sec:eval_HLA-Chat++_Friends}
Due to the time-consuming nature of annotating multi-party dialogue datasets and privacy issues, currently available open-source multi-party dialogue datasets are limited. Therefore, we primarily conducted evaluations on Ubuntu IRC-16 and Ubuntu IRC-19.
To verify the generalization ability of DRCR to other domains, we conducted tests on the HLA-Chat++~\cite{HLA-Chat++} and Friends~\cite{Friends} datasets, where HLA-Chat++ is a personalized multi-party dialogue dataset, and Friends includes the textual records from the TV show ``Friends''. The dataset divisions for HLA-Chat++ and Friends can be found in Table~\ref{tab_data}. Since HLA-Chat++ focuses on personalized conversations, in addition to using the BLEU metric to assess the accuracy of the generated responses, we also use the Dist-1/2 metric to measure the proportion of distinct unigrams and bigrams in the generated responses following previous work on personalized conversations. The higher the Dist-1/2 metric, the greater the diversity of responses generated by the model, and the better the personalization of those responses.

\begin{table}[t]
    \centering
    \resizebox{\linewidth}{!}{
    \begin{tabular}{lcccc}
        \toprule
        \multicolumn{1}{c}{\textbf{Method}} &  \bone & \btwo & D$_1$ & D$_2$  \\        
        \midrule
        HeterMPC  & 11.65	& 10.82	& 0.76	& 4.71  \\
        MADNet  & 12.39	& 11.84	& 0.94	& 5.49  \\
        SS-MPC  & 13.07	& 13.82	& 1.42	& 6.65  \\
        PersonaTKG+sent  & 13.18	& 14.17	& 1.59	& 6.90  \\
        Qwen3-8B+SFT & 13.42	& 14.48	& 1.72	& 7.04  \\
        \cellcolor[gray]{0.9}Qwen3-8B+DRCR (Ours) & \cellcolor[gray]{0.9}\textbf{15.83}	& \cellcolor[gray]{0.9}\textbf{17.06}	& \cellcolor[gray]{0.9}\textbf{3.53}	& \cellcolor[gray]{0.9}\textbf{9.23}  \\
        \bottomrule
    \end{tabular}
    }
    \caption{
    Performance evaluation of DRCR and baselines on the HLA-Chat++ dataset.
    }
    \label{table:evaluation_on_HLA-Chat++}
   \vspace{-3mm}
\end{table}

\begin{table}[t]
    \centering
    \resizebox{\linewidth}{!}{
    \begin{tabular}{lcccc}
        \toprule
        \multicolumn{1}{c}{\textbf{Method}} &  \bone & \btwo & \bthree & \rougel  \\        
        \midrule
        HeterMPC  & 5.30	& 1.41	& 0.40	& 4.74  \\
        MADNet  & 6.00	& 1.54	& 0.63	& 5.95  \\
        SS-MPC  & 7.81	& 2.42	& 1.47	& 7.24  \\
        Qwen3-8B+SFT & 7.13	& 2.06	& 1.39	& 6.82  \\
        \cellcolor[gray]{0.9}Qwen3-8B+DRCR (Ours) & \cellcolor[gray]{0.9}\textbf{9.27}	& \cellcolor[gray]{0.9}\textbf{5.14}	& \cellcolor[gray]{0.9}\textbf{2.83}	& \cellcolor[gray]{0.9}\textbf{9.56}  \\
        \bottomrule
    \end{tabular}
    }
    \caption{
    Performance evaluation of DRCR and baselines on the Friends dataset.
    }
    \label{table:evaluation_on_Friends}
    \vspace{-5mm}
\end{table}

The experimental results on HLA-Chat++ and Friends are shown in Tables~\ref{table:evaluation_on_HLA-Chat++} and~\ref{table:evaluation_on_Friends}, respectively. Compared to the baselines that directly perform supervised fine-tuning, the preference alignment and mutual self-evolution we designed have achieved further improvements, surpassing the baseliness in both accuracy and diversity. This also shows that DRCR can generalize to different domains (e.g., personalized conversations and daily conversations). Additionally, we also noticed that due to some topic shift phenomena in the Friends dataset, which brings difficulty to dialogue understanding, its performance is relatively low. How to design rewriting under multi-topic scenarios is also a direction worth exploring in the future.

\begin{table}[t]
      \centering
      \resizebox{\linewidth}{!}{
      \begin{tabular}{lccccc}
      \toprule
        Method               &   Coh.   &   Flu.   &  Info.  & Help.	&  Overall    \\
      \midrule
        Ground Truth           &   0.82   &  0.89    &  0.93  & 0.87  &   3.51   \\
      \midrule
        MADNet         &   0.58   &   0.87   & 0.76 &  0.63 &   2.84 \\
        RL-TRC         &   0.69   &   0.90  & 0.81 & 0.72 &   3.12 \\
        SS-MPC         &    0.73  &    0.91  & 0.82 &  0.79 &   3.25 \\
        Qwen3-8B+SFT         &   0.71   &  0.90    & 0.82  & 0.76 &   3.19  \\
        Qwen3-8B+DRCR         &   \textbf{0.79}   &   \textbf{0.94}   &  \textbf{0.86} &  \textbf{0.82}  &   \textbf{3.41}  \\
      \bottomrule
      \end{tabular}
      }
      \caption{Human evaluation on DRCR and four strong baselines. Coh., Flu., Info., and Help. are abbreviations for Coherence, Fluency, Informativeness, and Helpfulness, respectively.}
      \vspace{-6mm}
      \label{tab_human_eval}
\end{table}

\subsection{Human Evaluation}
We randomly sampled 200 samples from the test set and recruited three graduate students in NLP to score their responses based on coherence, fluency, informativeness, and helpfulness. Each evaluator assigns a binary score to each dimension. The evaluation results (average score) are shown in Table~\ref{tab_human_eval}. The results indicate that the agreement rate among the three evaluators reached 90\%, verifying the reliability of the human evaluation. We observed that DRCR consistently outperformed other methods across all four evaluation dimensions, achieving quality nearly on par with human-annotated responses. Notably, DRCR demonstrated the most substantial improvement in coherence, surpassing the best-performing baseline (\texttt{SS-MPC}) by 0.06. This is attributed to the model's integration of dialogue context rewriting informed by discourse coherence, which facilitates the generation of more contextually aligned responses. To further validate the performance of DRCR, we employed LLM for evaluation in Appendix~\ref{sec:llm_eval}.

\begin{figure}[t]
\begin{center}
 \includegraphics[width=1\linewidth]{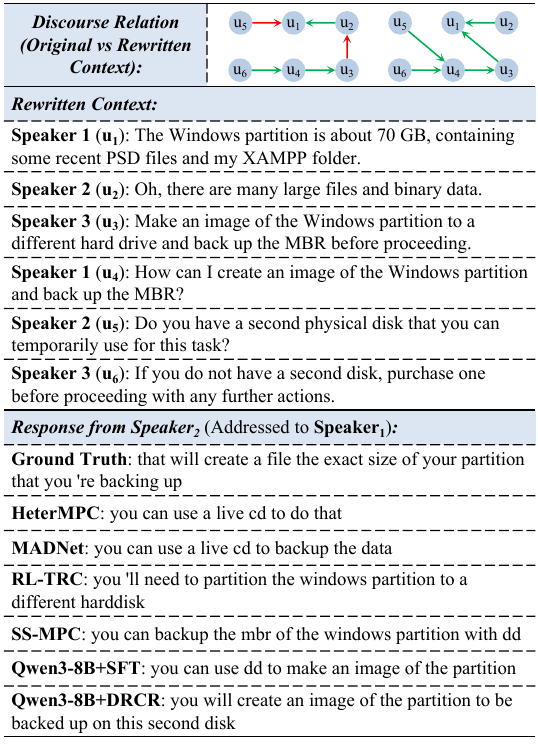}
 \caption{Case study on context rewriting and response generation. We provide the original context in Figure~\ref{fig:task}.}
 \label{fig:case}
\end{center}
 \vspace{-0.5cm}
\end{figure}

\subsection{Case Study}
As shown in Figure~\ref{fig:case} (with the original context referring to Figure~\ref{fig:task}), we conducted a case study on DRCR and five strong baselines (based on the original context). Before rewriting, the classifier mistakenly identified the addressees of $u_5$ and $u_3$ as $u_1$ and $u_2$. This was due to colloquial expressions and omissions in the dialogue that caused interference, thereby affecting the generation of response. After being rewritten by DRCR, the expression of $u_2$ is clearer, and the coreference resolution of ``that'' in $u_4$ has also been completed. 

We noticed that the responses generated by the other five baselines did not take into account that speaker 2 had already mentioned using a second disk in $u_5$, so they directly proposed a new solution in their responses (e.g., MADNet's response advocated using ``a live cd''), resulting in poor coherence and continuity with $u_5$. We provided a failed case in Appendix~\ref{sec:fc_analysis} for further insights into DRCR.

\section{Conclusion}
In this paper, we propose DRCR, which employs dialogue discourse coherence and response quality to guide dialogue context rewriting. By explicitly reformulating the dialogue context to enhance structural clarity and coherence, DRCR facilitates more effective multi-party dialogue generation. The experimental results on four datasets demonstrate that DRCR effectively improves the quality of response generation in multi-party dialogue.

\section*{Limitations}
Although DRCR can advance research on multi-party dialogue generation, there are two drawbacks. On one hand, although DRCR alleviates the reliance on external data through mutual self-evolution, it still requires a powerful LLM to construct preference data to initialize the behavior of rewriter and responder. On the other hand, rewriting introduces additional time overhead for some samples, as not every utterance in a conversation context needs to be rewritten. Future research can focus on how to selectively rewrite each utterance.

\section*{Acknowledgements}
The authors would like to thank the anonymous reviewers for their comments on this paper. This research was supported by the National Natural Science Foundation of China (Nos. 62276177 and 62376181), and Project Funded by the Priority Academic Program Development of Jiangsu Higher Education Institutions.

\bibliography{custom}

\newpage
\appendix

  \begin{table}[t]
    \centering
    \resizebox{\linewidth}{!}{
    \begin{tabular}{lcccccc}
        \toprule
        & \multicolumn{6}{c}{\textbf{Ubuntu IRC-19}} \\
         \cmidrule(lr){2-7}
        \multicolumn{1}{c}{\textbf{Variant}} &  \bone & \btwo & \bthree & \bfour & \meteor & \rougel \\        
        \midrule
        \cellcolor[gray]{0.9}DRCR & \cellcolor[gray]{0.9}17.81 & \cellcolor[gray]{0.9}8.02 & \cellcolor[gray]{0.9}5.79  & \cellcolor[gray]{0.9}4.26 & \cellcolor[gray]{0.9}8.49 & \cellcolor[gray]{0.9}14.82 \\
        \midrule
        DRCR (MPC)  & 17.63 & 7.86 & 5.68 & 4.18 & 8.36 & 14.73 \\
        DRCR (Acc) & 17.32 & 7.47 & 5.44 & 3.78 & 7.82 & 14.27  \\
        \bottomrule
    \end{tabular}
    }
    \caption{
    Analysis of using MPC-BERT as the addressee recognition classifier for discourse coherence scoring and using accuracy as the coherence proxy.
    }\label{table:discourse_coherence_scoring}
\end{table}

\section{Addressee Recognition Module}
\label{sec:Addressee_Recognition_Module}
Due to the limited existing research on addressee recognition, we implemented a lightweight addressee recognition classifier in this paper. Compared to the previous SOTA method MPC-BERT~\cite{MPC-BERT}, our method introduces biaffine attention for better relation modeling and removes other complex multi-task learning modules. In addition, compared to MPC-BERT using BERT~\cite{BERT} to encode dialogue context, we used the more advanced RoBERTa~\cite{RoBERTa} for encoding. 

We analyzed the effect of using MPC-BERT as a addressee recognition classifier for discourse coherence scoring, as shown in ``DRCR (MPC)'' in Table~\ref{table:discourse_coherence_scoring}. After using MPC-BERT, there is not much impact on the model's performance. This phenomenon occurs because our discourse coherence scoring module relies on the probabilities predicted by the model, which reflect the model's uncertainty. Although different methods may show variations in addressee recognition performance, their uncertainties can effectively reflect the differences in coherence after context rewriting. This is a relative change that does not depend on specific prediction results, reducing the bias in discourse coherence scoring caused by incorrect prediction results.

\section{Prompts in Rewriting and Responding}
\label{sec:utterance_level_rewriting}
We provide the prompts used for rewriting and response generation in Figure~\ref{fig:utterance_level_rewriting} and~\ref{fig:response_generation}, respectively. To generate a set of candidate rewritten utterances, we employed a sampling approach with varied temperature parameters.

\section{Implementation Details}
\label{sec:implementation}
In the importance calibration using the Coefficient of Variation, the temperature coefficient $\tau$ is set to 0.9. We use Low-Rank Adaptation (LoRA)~\cite{LoRA} for parameter-efficient training of the rewriter and responder. We set the LoRA rank and alpha parameters to 8 and 32, respectively. The number of iteration rounds for mutual self-evolution is set to 5, and the rewriter and responder are trained for two epochs in each iteration. The threshold $\phi$ in adaptive rewriting is set to 0.4. For the responder $\pi_\varepsilon$, we first perform supervised fine-tuning for 3 epochs on the original dataset. The statistics of the four datasets are shown in Table~\ref{tab_data}. 

\begin{table}[t]
      \centering
      \resizebox{\linewidth}{!}{
      \begin{tabular}{cccc}
      \toprule
        Dataset               &   Train   &   Valid   &  Test    \\
      \midrule
        Ubuntu IRC-16          &  461,120  &  28,570   &  32,668  \\
        Ubuntu IRC-19         &  311,725  &  5,000    &  5,000  \\
        HLA-Chat++         &  853,057  &  10,281    &  9,985  \\
        Friends         &  4,500  &  500    &  653  \\
      \bottomrule
      \end{tabular}
      }
      \caption{Statistics of the MDG datasets.}
      \label{tab_data}
\end{table}

\begin{table}[t]
    \centering
    \resizebox{\linewidth}{!}{
    \begin{tabular}{lcccccc}
        \toprule
        \multicolumn{1}{c}{\textbf{Method}} &  \bone & \btwo & \bthree & \bfour & \meteor & \rougel  \\        
        \midrule
        DRCR (GPT-3.5)	& 17.81	& 8.02	& 5.79	& 4.26	& 8.49	& 14.82  \\
        DRCR (GPT-4.1)	& 17.97	& 8.16	& 5.87	& 4.37	& 8.62	& 14.93  \\
        DRCR (Qwen3-4B)	& 16.53	& 6.74	& 4.92	& 3.26	& 7.31	& 13.74  \\
        DRCR (Qwen3-8B)	& 17.02	& 7.12	& 5.14	& 3.47	& 7.52	& 14.03  \\
        DRCR (Qwen3-32B)	& 18.13	& 8.32	& 5.98	& 4.52	& 8.79	& 15.12  \\
        \bottomrule
    \end{tabular}
    }
    \caption{
    Analysis of training data constructed using different LLMs on Ubuntu IRC-19.
    }
    \label{table:teacher_models_data_con}
\end{table}

\begin{table}[t]
    \centering
    \resizebox{\linewidth}{!}{
    \begin{tabular}{lcccccc}
        \toprule
        \multicolumn{1}{c}{\textbf{Method}} &  \bone & \btwo & \bthree & \bfour & \meteor & \rougel  \\        
        \midrule
        GPT-4.1 (zero-shot) & 15.37 & 7.16 & 4.35 & 2.91 & 7.13 & 13.49 \\
        GPT-4.1 (1-shot) & 15.54 & 7.42 & 4.55 & 3.06 & 7.35 & 13.72 \\
        GPT-4.1 (4-shot) & 15.46 & 7.28 & 4.31 & 2.97 & 7.22 & 13.56 \\
        Qwen3-32B (zero-shot) & 15.68 & 7.37 & 4.56 & 3.24 & 7.46 & 13.61 \\
        Qwen3-32B (1-shot) & 15.77 & 7.51 & 4.69 & 3.43 & 7.65 & 13.74 \\
        Qwen3-32B (4-shot) & 15.73 & 7.46 & 4.63 & 3.39 & 7.54 & 13.68 \\
        Qwen3-8B+DRCR (Ours)  & 17.81 & 8.02 & 5.79 & 4.26 & 8.49 & 14.82 \\
        \bottomrule
    \end{tabular}
    }
    \caption{
    Analysis of using powerful LLMs for multi-party dialogue generation on Ubuntu-19 without rewriting.
    }
    \label{table:wo_rewriting}
   \vspace{-5mm}
\end{table}

\section{Necessity of Context Rewriting}
\label{context_rewriting_necessity}

Our preliminary investigations suggest that even powerful LLMs still struggle with several characteristic challenges of multi-party dialogues:

\noindent \textbf{1. Interleaved thread disambiguation}. When multiple conversation threads are active simultaneously, even strong LLMs can conflate information across threads, producing responses that incorrectly merge topics or address the wrong conversational thread. This is because the flat sequential presentation of multi-party dialogue obscures the underlying reply structure, and even models with long context windows and strong reasoning capabilities are not immune to such structural ambiguity.

\noindent \textbf{2. Addressee and reference resolution}. Multi-party dialogues frequently contain implicit addressees and pronouns whose antecedents span across utterances from different speakers. While strong LLMs handle coreference well in well structured text, the fragmented and colloquial nature of multi-party conversation degrades their performance noticeably.

\noindent \textbf{3. Information scattering}. Relevant context for generating an appropriate response may be distributed across many non-adjacent utterances from different participants. Even when a strong LLM can in principle attend to all of them, the noisy intervening utterances from other threads can dilute attention and lead to incomplete or unfaithful responses.

In order to further analyze the importance of context rewriting for multi-party dialogue generation, we use zero-shot and few-shot methods to prompt LLMs to generate responses for multi-party dialogues without rewriting. Taking GPT-4.1 and Qwen3-32B as examples, the experimental results are shown in Table~\ref{table:wo_rewriting}. GPT-4.1 and Qwen3-32B perform worse than Qwen3-8B equipped with DRCR in zero-shot, 1-shot, and 4-shot scenarios. Additionally, we observed that increasing the number of examples did not lead to a significant performance improvement. This is potentially due to ambiguous expressions present in multi-party dialogues, which prevent the model from understanding these examples and result in generated responses deviating from the context.

\section{Coherence Assessment}
\label{sec:Coherence_Assessment}
In DRCR, coherence evaluation uses the probability of the ground truth predicted by the addressee classifier as a proxy. We also experimented with using the accuracy of the classifier prediction directly as the coherence score, as shown in ``DRCR (Acc)'' in Table~\ref{table:discourse_coherence_scoring}. Using accuracy to measure coherence is less effective than using probabilities, which can be attributed to the fact that the sum of probabilities can be seen as a soft form of accuracy, whereas directly using accuracy cannot reflect such fine-grained information, which encompasses the uncertainty of the classifiers.

\section{Different Teacher Models for Data Construction}
\label{sec:teacher_models_data_con}
In addition to the analysis using GPT-4.1 and Qwen3-8B as teacher models, we further conducted tests on LLMs of other scales. As shown in Table~\ref{table:teacher_models_data_con}, we evaluated the training data constructed using \texttt{Qwen3-4B} and \texttt{Qwen3-32B} respectively. For all teacher models, we use the original Ubuntu IRC-16 and IRC-19 dataset as seed data for distillation.

First, we can observe that the models trained on the data constructed by stronger foundation models exhibit better performance. This is attributed to the higher quality of the generated data, resulting in more powerful rewriter and responder after training. 

\begin{figure}[t]
\begin{center}
 \includegraphics[width=1\linewidth]{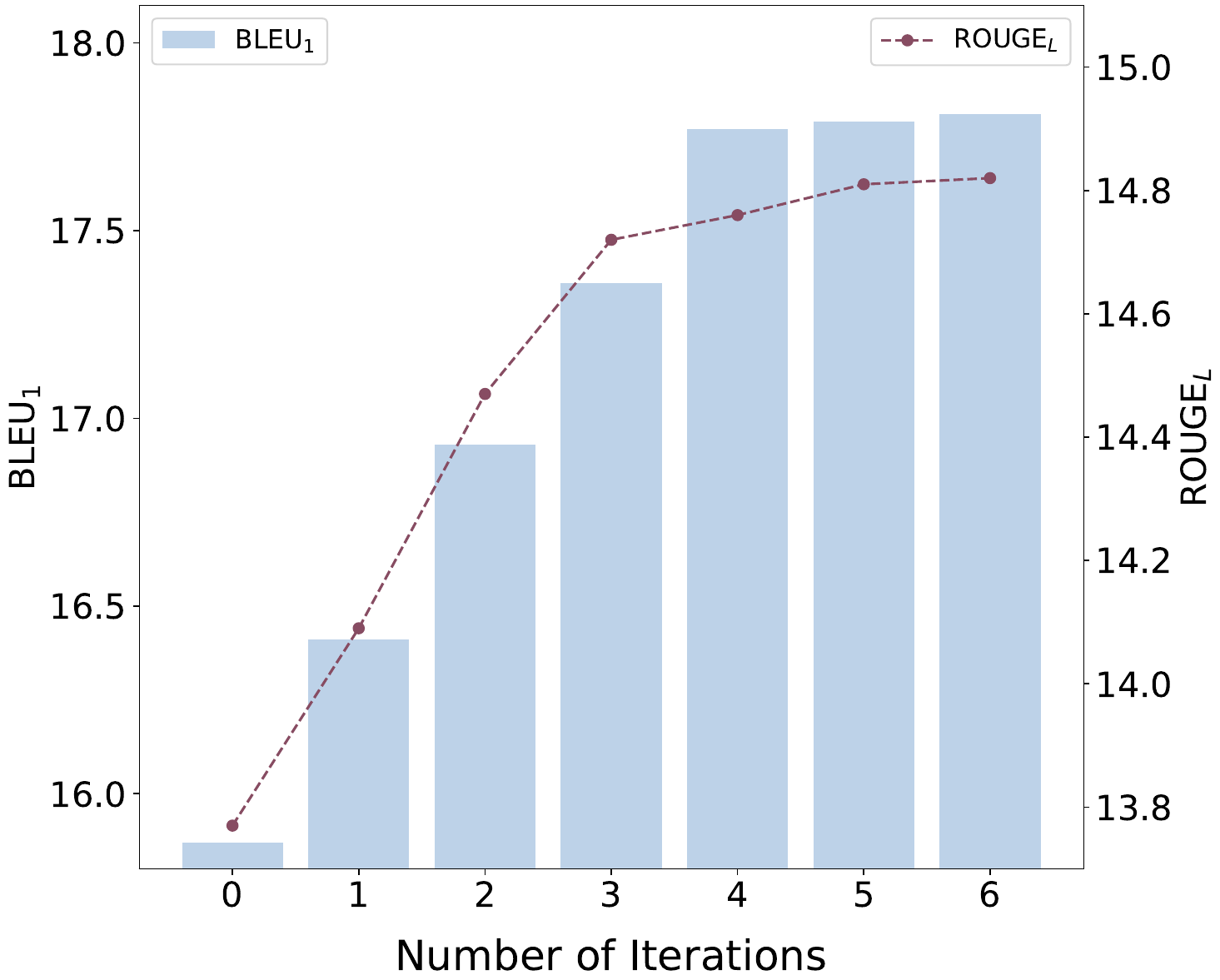}
 \caption{Analysis of scaling self-evolution rounds on Ubuntu IRC-19.}
 \label{fig:scale_iterations}
\end{center}
\end{figure}

On the other hand, although the capability of the teacher model affects the ability of the student model, which is similar to the finding of knowledge distillation, the gap between the influence of different teacher models on student model is relatively small. Compared to \texttt{Qwen3-8B} as the teacher model, \texttt{Qwen3-4B} as the teacher model only decreases by 0.49 on the BLEU1 metric. The performance of Qwen3-8B on BLEU1 with only supervised fine-tuning (i.e., ``Qwen3-8B+SFT'' In Table \ref{table:main_exp}) is only 13.07, which is much worse than the performance of \texttt{Qwen3-4B} as a teacher model.

The performance fluctuations due to different LLMs on the data construction of the warm-up phase are minor compared to the enhancements brought about by mutual self-evolution (e.g., the $\text{B}_\text{1}$ on IRC-19 improved by 1.94 in Table~\ref{table:ablation}). This is due to the fact that the warm-up phase serves to initialize the rewriter and responder, which can continue to enhance each other during mutual self-evolution.

\section{Iteration Number in Mutual Self-Evolution}
\label{sec:scale_iterations}
In the process of mutual self-evolution, we set 5 iterations by default. Taking Ubuntu IRC-19 as an example, we analyzed the changes in the process of scaling the iteration rounds from 1 to 6. As shown in Figure~\ref{fig:scale_iterations}, with the increase in self-mutual evolution iteration rounds, the model performance continuously improves and gradually saturates after the fourth round. In this process, the rewriter and responder improved performance through iterative optimization of preference data and preference alignment.

\section{Candidate Number in Mutual Self-Evolution}
\label{sec:scale_candidates}

\begin{figure}[t]
\begin{center}
 \includegraphics[width=1\linewidth]{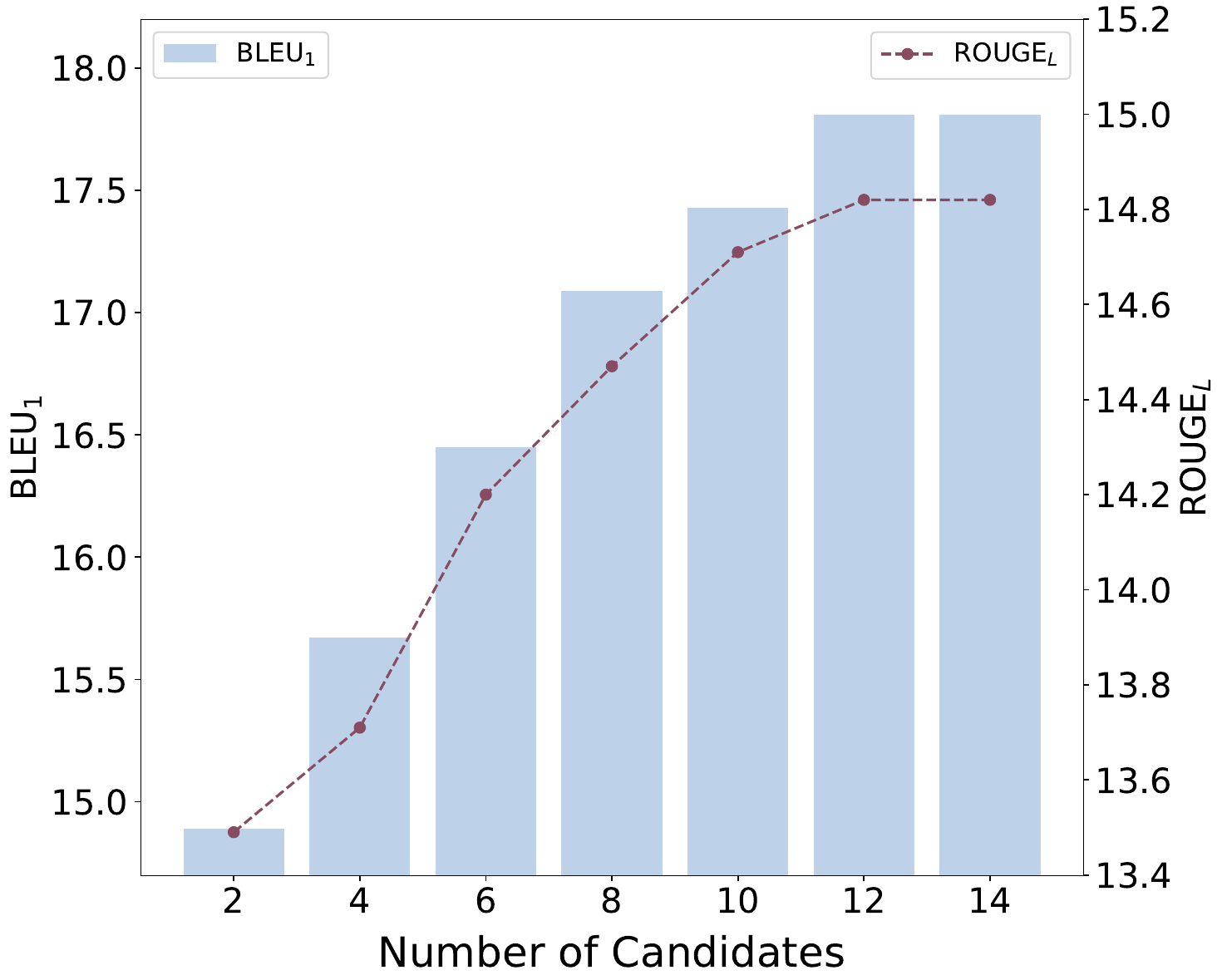}
 \caption{Analysis of scaling sampled candidate contexts on Ubuntu IRC-19.}
 \label{fig:scale_candidates}
\end{center}
\end{figure}

\begin{figure}[t]
\begin{center}
 \includegraphics[width=1\linewidth]{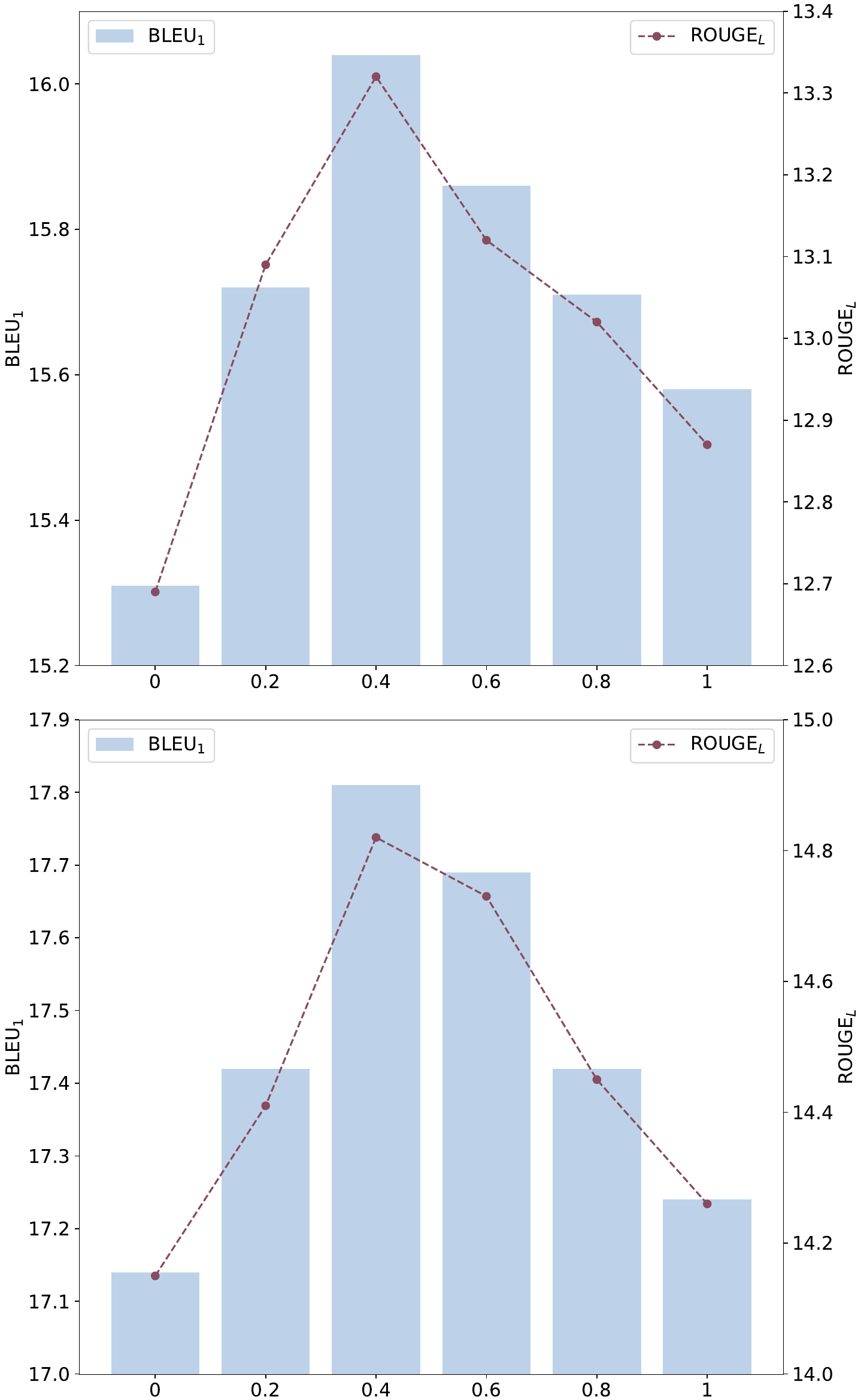}
 \caption{The impact of rewriting threshold $\phi$ on model performance in Ubuntu IRC-16 (top) and Ubuntu IRC-19 (bottom).}
 \label{fig:rewriting_threshold}
\end{center}
\end{figure}
\begin{table}[t]
    \centering
    \resizebox{\linewidth}{!}{
    \begin{tabular}{lcc}
        \toprule
        \multicolumn{1}{c}{\textbf{Stage}} &  Ubuntu IRC-16 & Ubuntu IRC-19  \\        
        \midrule
        Data Construction  & 13.2	& 18.7  \\
        Warm-up Training & 7.4	& 11.8	  \\
        Mutual Self-evolution	& 12.1	& 17.2	  \\
        \bottomrule
    \end{tabular}
    }
    \caption{
    Analysis of time consumption TC (in hours) in the data construction and training phase.
    }
    \label{table:time_consumption}
\end{table}

\begin{table}[t]
    \centering
    \resizebox{\linewidth}{!}{
    \begin{tabular}{lcc}
        \toprule
        \multicolumn{1}{c}{\textbf{Variant}} &  Ubuntu IRC-16 & Ubuntu IRC-19  \\        
        \midrule
        Non-rewriting  & 0.37	& 0.42  \\
        Adaptive Rewriting & 0.58	& 0.71	  \\
        \bottomrule
    \end{tabular}
    }
    \caption{
    Taking Qwen3-8B as an example to analyze the inference latency PL (in seconds) of non-rewriting and adaptive rewriting.
    }
    \label{table:inference_latency}
\end{table}
During the mutual self-evolution of the rewriter and responder, DRCR samples $m$ candidate rewritten contexts and their corresponding responses to construct preference data after each iteration. This process can be regarded as a self-exploration process. We set different values of $m$ to analyze the impact of exploration space on model performance. As shown in Figure~\ref{fig:scale_candidates}, increasing the number of sampled candidate contexts within a certain range can improve model performance. This is due to the availability of higher quality preference data after expanding the exploration space, which in turn facilitates the training of the rewriter and responder.

\section{Analysis of Rewriting Threshold $\phi$}
\label{sec:rewriting_threshold}
During the rewriting process, we adaptively rewrite the dialogue context by setting the threshold $\phi$. We analyzed the effect of setting different thresholds on model performance. Specifically, when $\phi=0$, no rewriting is performed, and when $\phi=1$, the dialogue context of all samples is rewritten. As shown in Figure ~\ref{fig:rewriting_threshold}, with the increase of the rewriting threshold, the model's performance first rises and then falls. This phenomenon occurs because a lower rewriting threshold introduces noise to the model's understanding due to some informal and ambiguous contexts. However, the rewritten context may have deviations compared to the original context, so when the threshold is larger, incorrect rewriting interferes with the model's predictions.

\section{Time Complexity and Resource Usage}
\label{sec:time_complexity_resource_usage}
Since tree-based sampling and mutual self-evolution are performed in the data construction and training phases respectively, they are one-time processes and will not increase the inference latency, thus not affecting the deployment in the real world. In the inference phase, we use an adaptive rewriting approach, i.e., rewriting selectively based on the coherence of the dialogue history. We report the Time Consumption (TC) during the data construction phase, warm-up training phase, and mutual self-evolution training phase in Table~\ref{table:time_consumption}. In Table~\ref{table:inference_latency}, we report the Per-sample Latency (PL) caused by adaptive rewriting and non-rewriting. All our experiments were conducted on a single NVIDIA A100 PCIe GPU with 40GB of VRAM.

We can observe that the data construction and training on a single dataset can be completed within two days on a single A100 PCIe GPU, and acceleration can be achieved by using multi-GPU training. In the case of adaptive rewriting, only a small number of samples need to be rewritten, and the average latency on Ubuntu IRC-16 and Ubuntu IRC-19 increases by only 0.21 and 0.29 seconds compared to non-rewriting, which is acceptable for humans.

\section{LLM-as-a-Judge for Response Evaluation}
\label{sec:llm_eval}
To further verify the effectiveness of DRCR, we sampled 200 pieces of data from the test set and used GPT-4 as an evaluator to score the responses generated by DRCR and baselines in four dimensions: coherence, fluency, informativeness, and helpfulness. The prompt we used is shown in Figure~\ref{fig:llm_eval}.

The experimental results are shown in the Table~\ref{table:llm_as_a_judge}. ``Overall'' denotes the average score of the four dimensions. DRCR outperforms the previous baselines and the direct supervised fine-tuning Qwen3-8B+SFT in all four dimensions of evaluation. This is attributed to DRCR's ability of guiding the model in understanding the semantics of dialogue history through rewriting, thereby generating higher quality responses.

\begin{table}[t]
    \centering
    \resizebox{\linewidth}{!}{
    \begin{tabular}{lccccc}
        \toprule
        \multicolumn{1}{c}{\textbf{Method}} &  Coh.	& Flu.	& Info.	& Help.	& Overall  \\        
        \midrule
        Ground Truth  & 3.67	&  4.42	&  3.81	&  4.32	&  4.06  \\
        \midrule
        MADNet & 2.85	&  3.64	&  3.19	&  3.37	&  3.26  \\
        RL-TRC & 3.01	&  3.89	&  3.40	&  3.62	&  3.48  \\
        SS-MPC & 3.18	&  4.12	&  3.53	&  3.87	&  3.68  \\
        Qwen3-8B+SFT & 3.12	&  4.03	&  3.51	&  3.83	&  3.62  \\
        Qwen3-8B+DRCR & 3.41	&  4.36	&  3.87	&  4.12	&  3.94  \\
        \bottomrule
    \end{tabular}
    }
    \caption{
    The generated responses are evaluated from four dimensions: coherence, fluency, informativeness, and helpfulness using GPT-4. Coh., Flu., Info., and Help. are abbreviations for coherence, fluency, informativeness, and helpfulness, respectively.
    }
    \label{table:llm_as_a_judge}
\end{table}

\section{Analysis of Failure Case}
\label{sec:fc_analysis}
Due to the noise information in the context of the dialogue, the rewritten utterances may deviate from the original utterances in terms of semantics. We provide an example in Figure~\ref{fig:failure_case}. In this case, $u_1$ to $u_6$ are the original dialogue utterances and $v_1$ to $v_6$ are the rewritten dialogue utterances. The model needs to predict speaker 2's response to $u_5$. There are two dialogue flows involved in this dialogue, $u_1 \xrightarrow{} u_2 \xrightarrow{} u_6$ and $u_3 \xrightarrow{} u_4 \xrightarrow{} u_5$, which are coupled together. However, we notice that the rewriter mixes the two dialogue flows together, e.g., ``hardware-driver tool'' is introduced in $v_6$ in the dialogue after the rewrite, yet what speaker 4 is trying to express is that the lenovo x41 can be used without the touch-screen driver instead of the hardware-driver tool. Such rewriting subsequently led to the generated response being interfered with by $v_6$, mistakenly understood as needing to enable restricted drivers on the Lenovo X41. However, the response and $v_6$ are not in the same dialogue flow.

Future research can focus on how to guide models in rewriting, reducing hallucinations during the rewriting process. For example, extracting events from dialogues to assist models in rewriting dialogue context.

\begin{figure*}
    \centering
   \begin{tcolorbox}[
      colback=white, colframe=black, arc=3mm, width=\linewidth,
      title=\textbf{Prompt for Utterance-level Rewriting}, 
      coltitle=white, colbacktitle=gray, fonttitle=\bfseries
    ]
    You will be provided with a conversation among multiple people. The last utterance in the conversation may be expressed in a less formal way, or there may be co-references or omissions. Please rewrite the last utterance, such as performing coreference resolution and ellipsis resolution, to make its expression more formal and complete without introducing redundant information. Only output the rewritten last utterance without additional content.
    
    \vspace{2pt}
    
    Here are some examples.
    
    \textbf{Example 1}:
    
    \textbf{Conversation}:
    
    Speaker 1: oh my god my internet is soooo slooooowwww
    
    \textbf{Rewritten Utterance}:
    
    My internet is so slow.

    \vspace{12pt}

    \textbf{Example 2}:
    
    \textbf{Conversation}:
    
    Speaker 1: I think it is because of permissions.
    
    Speaker 2: Why are you using an RPM package in Ubuntu?
    
    Speaker 2: You should use sudo to execute commands with superuser privileges.
    
    Speaker 2: Additionally, Ubuntu uses Debian (deb) packages, not Red Hat Package Manager (rpm) packages.
    
    Speaker 3: just put it in front of any command that you want to run as root .
    
    \textbf{Rewritten Utterance}:
    
    Just put ``sudo'' in front of any command that you wish to execute with root privileges.

    \vspace{12pt}
    
    \textbf{Example 3}:
    
    \textbf{Conversation}:
    
    Speaker 1: is ubuntu 7.01 is compatilable with windows vista ? ? ? ?
    
    Speaker 2: compatible in what way ? you can have both 7.10 and vista on same system
    
    Speaker 3: do you mean : can the grub bool loader boot windows vista ?
    
    Speaker 4: the only compatibility issue is with it 's hardware
    
    \textbf{Rewritten Utterance}:
    
    The only compatibility issue between Ubuntu 7.10 and Windows Vista pertains to the hardware requirements of the system.

    \vspace{12pt}
    
    Please rewrite the last utterance of the following conversation. The rewritten utterances need to be as concise as possible, retaining important content, and not exceeding 20 words per utterance after rewriting.
    
    \textbf{Conversation}:
    
    \textcolor{darkgreen}{\{Conversation\_context\}}
    
    \textbf{Rewritten Utterance}:

    \end{tcolorbox}
    \caption{The prompt used for utterance-level rewriting in Section~\ref{PDC}.}
    \label{fig:utterance_level_rewriting}
\end{figure*}

\begin{figure*}
    \centering
   \begin{tcolorbox}[
      colback=white, colframe=black, arc=3mm, width=\linewidth,
      title=\textbf{Prompt for Response Generation}, 
      coltitle=white, colbacktitle=gray, fonttitle=\bfseries
    ]
    You will be provided with a conversation among multiple people. The number of the utterance being replied to in the current round of dialogue is provided at the beginning of each round, and this number is placed in square brackets.
    
    \vspace{2pt}
    
    Here are some examples.
    
    \textbf{Example 1}:
    
    \textbf{Conversation}:
    
    [1] Speaker 1: what is the best desktop search for ubuntu ? i just found beagle

    [1] Speaker 2: best is subjective , but tracker was included by default in gutsy , so i suppose you could say that ubuntu developers think tracker is the best

    [2] Speaker 3: so stop the whining and use masm

    [1] Speaker 4: find -name 'keyword ' for the win

    [1] Speaker 2: tracker ? it 's a desktop search applications .

    [3] Speaker 2: then how , pray tell , will i be using it ?
    
    [3] Speaker 2: 
    
    \textbf{Response}:
    
    is there some magical linux port of masm that i have n't heard of ?

    \vspace{12pt}
    
    \textbf{Example 2}:
    
    \textbf{Conversation}:
    
    [1] Speaker 1: how can i conveniently open an iso ? like without mounting it from cli
    
    [1] Speaker 2: open it with archive manager
    
    [2] Speaker 1: does that work ? what archive manager= ?
    
    [3] Speaker 3: you can open it with the archive manager ( its name is `` file-roller '' , the app that opens FILEPATH files ) then it 'll be mounted as an archive , but you 'll have to extract the data from it . it 's not as convenient as mounting

    [3] Speaker 3: thus , install gmountiso if you want to graphically manage your mounting points EMOJI

    [3] Speaker 2: 
    
    \textbf{Response}:
    
    i just opened an iso with the built in archive manager

    \vspace{12pt}
    
    Please generate the final response based on the context and structure of the conversation. You only need to generate the response, do not output any extra content.
    
    \textbf{Conversation}:
    
    \textcolor{darkgreen}{\{Conversation\_context\}}
    
    \textbf{Response}:

    \end{tcolorbox}
    \caption{The prompt used for response generation in Section~\ref{PDC}.}
    \label{fig:response_generation}
\end{figure*}

\begin{figure*}
    \centering
   \begin{tcolorbox}[
      colback=white, colframe=black, arc=3mm, width=\linewidth,
      title=\textbf{Prompts Used for Response Evaluation by LLM}, 
      coltitle=white, colbacktitle=gray, fonttitle=\bfseries
    ]
    Different from a two-person conversation, a multi-party conversation involves multiple speakers. Now given a history of a multi-party conversation, the current response, and the addressee of each utterance, please score the current response from four dimensions: coherence, fluency, informativeness, and helpfulness, with a score range from 1 to 5. The explanations for these four dimensions are as follows.

    \vspace{12pt}

    Coherence: refers to how smoothly utterances are connected logically and semantically, and whether an utterance naturally follows on from the previous utterance to form a meaningful whole.

    \vspace{12pt}
    
    Fluency: refers to whether the language itself is natural, smooth and consistent with human expression. It is concerned with linguistic form rather than logical content.
    
    \vspace{12pt}
    
    Informativeness: the amount of valid, relevant and non-redundant information contained in a response. A highly informative response should be able to convey meaningful information.

    \vspace{12pt}
    
    Helpfulness: whether the response actually solves the addressee's problem or fulfills the addressee's need.

    \vspace{12pt}
    
    Your output format is a dictionary containing four items, each with a key as the name of the above four dimensions and a value as the corresponding score. Please do not output extra content.

    \end{tcolorbox}
    \caption{Prompt used to evaluate the quality of responses using LLM.}
    \label{fig:llm_eval}
\end{figure*}

\begin{figure*}
    \centering
   \begin{tcolorbox}[
      colback=white, colframe=black, arc=3mm, width=\linewidth,
      title=\textbf{A Failed Case Caused by Rewriting Error}, 
      coltitle=white, colbacktitle=gray, fonttitle=\bfseries
    ]
    Original Dialogue:
    
    Speaker 1 ($u_1$): can someone tell me where i can get a touchscreen driver for my tablet pc ?
    
    Speaker 2 ($u_2$): i would check the forums for your exact tablet pc and see what others are using .
    
    Speaker 3 ($u_3$): where is the hardware-driver tool ?
    
    Speaker 2 ($u_4$): in the system - admin menus
    
    Speaker 3 ($u_5$): i ran hardware-driver tool but it didnt find anything
    
    Speaker 4 ($u_6$): my lenovo x41 worked out of the box

    \vspace{12pt}
    
    Rewritten Dialogue:
    
    Speaker 1 ($v_1$): Can someone inform me where I can obtain a touchscreen driver for my tablet computer?
    
    Speaker 2 ($v_2$): I would recommend checking the forums specific to your tablet PC to see what other users have found useful or recommended.
    
    Speaker 3 ($v_3$): Where can I locate the hardware driver tool?
    
    Speaker 2 ($v_4$): The hardware-driver tool can be found in the system administration menus.
    
    Speaker 3 ($v_5$): I executed the hardware-driver tool, but it did not detect any drivers.
    
    Speaker 4 ($v_6$): My Lenovo X41 worked out of the box without requiring any hardware-driver tool.

    \vspace{12pt}
    
    Speaker 2's response to $u_5$:
    
    \textbf{Ground Truth}: no idea then . could be your card is one thats not supported by the driuvers yet . or some other oddity .

    \textbf{DRCR}: you may have to enable the restricted drivers in lenovo x41 .

    \end{tcolorbox}
    \caption{A case where incorrect rewriting affects response generation.}
    \label{fig:failure_case}
\end{figure*}

\end{document}